# Which Spatial Partition Trees are Adaptive to Intrinsic Dimension?

**Nakul Verma**
UC San Diego
naverma@cs.ucsd.edu

**Samory Kpotufe**
UC San Diego
skpotufe@cs.ucsd.edu

**Sanjoy Dasgupta**
UC San Diego
dasgupta@cs.ucsd.edu

## Abstract

Recent theory work has found that a special type of spatial partition tree – called a *random projection tree* – is adaptive to the intrinsic dimension of the data from which it is built. Here we examine this same question, with a combination of theory and experiments, for a broader class of trees that includes $k$-d trees, dyadic trees, and PCA trees. Our motivation is to get a feel for (i) the kind of intrinsic low dimensional structure that can be empirically verified, (ii) the extent to which a spatial partition can exploit such structure, and (iii) the implications for standard statistical tasks such as regression, vector quantization, and nearest neighbor search.

## 1 INTRODUCTION

A *spatial partitioning tree* recursively divides space into increasingly fine partitions. The most popular such data structure is probably the $k$-d tree, which splits the input space into two cells, then four, then eight, and so on, all with axis-parallel cuts. Each resulting partitioning has cells that are axis-aligned hyperrectangles (see Figure 1, middle). Once such a hierarchical partitioning is built from a data set, it can be used for standard statistical tasks. When a new query point $q$ arrives, that point can quickly be moved down the tree to a leaf cell (call it $C$). For classification, the majority label of the data points in $C$ can be returned. For regression, it will be the average of the response values in $C$. For nearest neighbor search, the closest point to $q$ in $C$ can be returned; of course, this might not be $q$'s nearest neighbor overall, but if the cell $C$ is sufficiently small, then it will at any rate be a point close enough to $q$ to have similar properties.

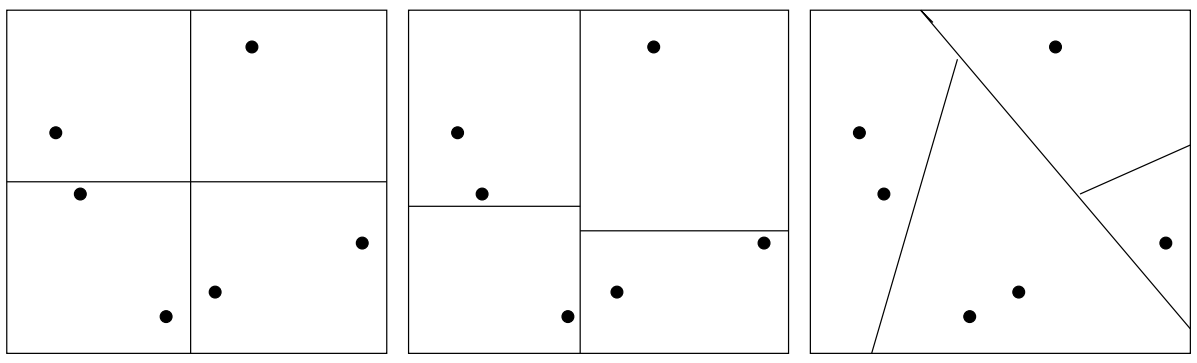

Figure 1: Examples of Spatial Trees. Left: dyadic tree – cycles through coordinates and splits the data at the mid point. Middle: $k$-d tree – picks the coordinate direction with maximum spread and splits the data at the median value. Right: RP tree – picks a random direction from the unit sphere and split the data at the median value.

There are different ways to build a $k$-d tree, depending on which split coordinate is chosen at each stage. There are also many other types of spatial partitioning trees, such as dyadic trees (Figure 1, left) and PCA trees. We are interested in understanding the relative merits of these different data structures, to help choose between them. A natural first step, therefore, is to look at the underlying statistical theory. This theory, nicely summarized in Chapter 20 of [DGL96], says that the convergence properties of tree-based estimators can be characterized in terms of the rate at which cell diameters shrink as you move down the tree. The more rapidly these cells shrink, the better.

For $k$-d trees, these cell diameters can shrink very slowly when the data is high-dimensional. For $D$-dimensional data, it may require $D$ levels of the tree (and thus at least $2^D$ data points) to just *halve* the diameter. Thus $k$-d trees suffer from the same curse of dimensionality as other nonparametric statistical methods. But what if the data has low intrinsic dimension; for instance, if it lies close to a low-dimensional manifold? We are interested in understanding the behavior of spatial partitioning trees in such situations.

Some recent work [DF08] provides new insights into this problem. It begins by observing that there is more than one way to define a cell's diameter. The statistical theory has generally considered the diameter to be the distance between the furthest pair of points on its boundary (if it is convex, then this is the distance between the furthest pair of vertices of the cell). It is very difficult to get bounds on this diameter unless the cells are of highly regular shape, such as hyperrectangles. A different, more flexible, notion of diameter looks at the furthest pair of *data points* within the cell, or even better, the typical interpoint distance of data within the cell (see Figure 3). It turns out that rates of convergence for statistical estimators can be given in terms of these kinds of *data diameter* (specifically, in terms of the rate at which these diameters decrease down the tree). Moreover, these data diameters can be bounded even if the cells are of unusual shapes. This immediately opens the door to analyzing spatial partitioning trees that produce non-rectangular cells.

[DF08] introduced *random projection trees* – in which the split at each stage is at the median along a direction chosen at random from the surface of the unit sphere (Figure 1, right) – and showed that the data diameter of the cells decreases at a rate that depends only on the *intrinsic dimension* of the data, not $D$:

> Let $d$ be the intrinsic dimension of data falling in a particular cell $C$ of an RP tree. Then all cells $O(d)$ levels below $C$ have at most half the data diameter of $C$.

(There is no dependence on the ambient dimension $D$.) They proved this for two notions of dimension: *Assouad dimension*, which is standard in the literature on analysis on metric spaces, and a new notion called *local covariance dimension*, which means simply that small enough neighborhoods of the data set have covariance matrices that are concentrated along just a few eigendirections.

We are interested in exploring these phenomena more broadly, and for other types of trees. We start by examining the notion of local covariance dimension, and contrast it with other notions of dimension through a series of inclusion results. To get more intuition, we then investigate a variety of data sets and examine the extent to which these data verifiably have low local covariance dimension. The results suggest that this notion is quite reasonable and is of practical use. We then consider a variety of spatial partition trees: (i) $k$-d trees (of two types), (ii) dyadic trees, (iii) random projection trees, (iv) PCA trees, and (v) 2-means trees. We give upper and lower bounds on the diameter decrease rates achieved by these trees, as a function of local covariance dimension. Our strongest upper bounds on these rates are for PCA trees and 2-means trees, followed by RP trees. On the other hand, dyadic trees and $k$-d trees are weaker in their adaptivity. Our next step is to examine these effects experimentally, again on a range of data sets. We also investigate how the diameter decrease rate is correlated with performance in standard statistical tasks like regression and nearest neighbor search.

## 2 INTRINSIC DIMENSION

Let $\mathcal{X}$ denote the space in which data lie. In this paper, we'll assume $\mathcal{X}$ is a subset of $\mathbb{R}^D$, and that the metric of interest is Euclidean ($L_2$) distance. How can we characterize the intrinsic dimension of $\mathcal{X}$? This question has aroused keen interest in many different scientific communities, and has given rise to a variety of definitions. Here are four of the most successful such notions, arranged in decreasing order of generality:

- Covering dimension
- Assouad dimension
- Manifold dimension
- Affine dimension

The most general is the *covering dimension*: the smallest $d$ for which there is a constant $C > 0$ such that for any $\epsilon > 0$, $\mathcal{X}$ has an $\epsilon$-cover of size $C(1/\epsilon)^d$. This notion lies at the heart of much of empirical process theory. Although it permits many kinds of analysis and is wonderfully general, for our purposes it falls short on one count: for nonparametric estimators, we need small covering numbers for $\mathcal{X}$, but also for individual *neighborhoods* of $\mathcal{X}$. Thus we would like this same covering condition (with the same constant $C$) to hold for all $L_2$-balls in $\mathcal{X}$. This additional stipulation yields the *Assouad dimension*, which is defined as the smallest $d$ such that for any (Euclidean) ball $B \subset \mathbb{R}^D$, $X \cap B$ can be covered by $2^d$ balls of half the radius.

At the bottom end of the spectrum is the *affine dimension*, which is simply the smallest $d$ such that $\mathcal{X}$ is contained in a $d$-dimensional affine subspace of $\mathbb{R}^D$. It is a tall order to expect this to be smaller than $D$, although we may hope that $\mathcal{X}$ lies close to such a subspace. A more general hope is that $\mathcal{X}$ lies on (or close to) a $d$-dimensional Riemannian submanifold of $\mathbb{R}^D$. This notion makes a lot of intuitive sense, but in order for it to be useful either in algorithmic analysis or in estimating dimension, it is necessary to place conditions on the curvature of the manifold. [NSW06] has recently suggested a clean formulation in which the curvature is captured by a single value which they call the *condition number* of the manifold. Similar notions have earlier been used in the computational geometry literature [AB98].

In what sense is our list arranged by decreasing generality? If $\mathcal{X}$ has an affine dimension of $d$, it certainly has manifold dimension at most $d$ (whatever the restriction on curvature). Similarly, low Assouad dimension implies small covering numbers. The only nontrivial containment result is that if $\mathcal{X}$ is a $d$-dimensional Riemannian submanifold with bounded curvature, then sufficiently small neighborhoods of $\mathcal{X}$ (where this neighborhood radius depends on the curvature) have Assouad dimension $O(d)$. This result is formalized and proved in [DF08]. The containment is strict: there is a substantial gap between manifolds of bounded curvature and sets of low Assouad dimension, on account of the smoothness properties of the former. This divide is not just a technicality but has important algorithmic implications. For instance, a variant of the Johnson Lindenstrauss lemma states that when a $d$-dimensional manifold (of bounded curvature) is projected onto a random subspace of dimension $O(d/\epsilon^2)$, then all interpoint distances are preserved within $1 \pm \epsilon$ [BW07], [Cla07]. This does not hold for sets of Assouad dimension $d$ [IN07].

None of these four notions arose in the context of data analysis, and it is not clear that any of them is well-suited to the dual purpose of (i) capturing a type of intrinsic structure that holds (verifiably) for many data sets and (ii) providing a formalism in which to analyze statistical procedures. In addition, they all describe sets, whereas in statistical contexts we are more interested in characterizing the dimension of a probability distribution. The recent machine learning literature, while appealing to the manifold idea for intuition, seems gradually to be moving towards a notion of "local flatness". [DF08] formalized this notion and called it the *local covariance dimension.*

## 2.1 LOCAL COVARIANCE DIMENSION

**Definition 1.** *Let $\mu$ be any measure over $\mathbb{R}^D$ and let $S$ be its covariance matrix. We say that $\mu$ has covariance dimension $(d, \epsilon)$ if the largest $d$ eigenvalues of $S$ account for $(1-\epsilon)$ fraction of its trace. That is, if the eigenvalues of $S$ are $\lambda_1 \geq \lambda_2 \geq \cdots \geq \lambda_D$, then*

$$\lambda_1 + \cdots + \lambda_d \geq (1-\epsilon)(\lambda_1 + \cdots + \lambda_D).$$

A distribution has covariance dimension $(d, \epsilon)$ if all but an $\epsilon$ fraction of its variance is concentrated in a $d$-dimensional affine subspace. Equivalently, the projection of the distribution onto this subspace leads to at most an $\epsilon$ total loss in squared distances. It is, in general, too much to hope that an entire data distribution would have low covariance dimension. But we might hope that this property holds *locally*; or more precisely, that all (or most) sufficiently-small neighborhoods have low covariance dimension. At this stage, we could make this definition more complicated by quantifying the "most" or "sufficiently small" (as [DF08] did to some extent), but it will turn out that we don't need to do this in order to state our theorems, so we leave things as they are.

Intuitively, the local covariance condition lies somewhere between manifold dimension and Assouad dimension, although it is more general in that it merely requires points to be close to a locally flat set, rather than exactly on it.

## 2.2 EXPERIMENTS WITH DIMENSION

Covariance dimension is an intuitive notion, and recalls standard constructs in statistics such as mixtures of factor analyzers. It is instructive to see how it might be estimated from samples, and whether there is evidence that many data sets do exhibit low covariance dimension.

First let's set our expectations properly. Even if data truly lies near a low-dimensional manifold structure, this property would only be apparent at a certain *scale*, that is, when considering neighborhoods whose radii lie within an appropriate range. For larger neighborhoods, the data set might seem slightly higher dimensional: the union of a slew of local low-dimensional subspaces. And for smaller neighborhoods, all we would see is pure noise, and the data set would seem full-dimensional.

Thus we will empirically estimate covariance dimension at different resolutions. First, we determine the diameter $\Delta$ of the dataset $\mathbf{X}$ by computing the maximum interpoint distance, and we choose multiple values $r \in [0, \Delta]$ as our different scales (radii). For each such radius $r$, and each data point point $x \in \mathbf{X}$, we compute the covariance matrix of the data points lying in the ball $B(x, r)$, and we determine (using a standard eigenvalue computation) how many dimensions suffice for capturing a $(1-\epsilon)$ fraction of the variance. In our experiments, we try $\epsilon = 0.1$ and $0.01$. We then take the dimension at scale $r$ (call it $d(r)$) to be average of all these values (over $x$).

How can we ascertain that our estimate $d(r)$ is indicative of the underlying covariance dimension at resolution $r$? If the balls $B(x, r)$ are so small as to contain very few data points, then the estimate $d(r)$ is not reliable. Thus we also keep track of $n(r)$, the average number of data points within the balls $B(x, r)$ (averaged over $x$). Roughly, we can expect $d(r)$ to be a reliable estimate if $n(r)$ is an order of magnitude larger than $d(r)$.

Figure 2 plots $d(r)$ against $r$ for several data sets. The numerical annotations on each curve represent the values $n(r)$. Loosely speaking, the larger the ratio $n(r)/d(r)$, the higher our confidence in the estimate.

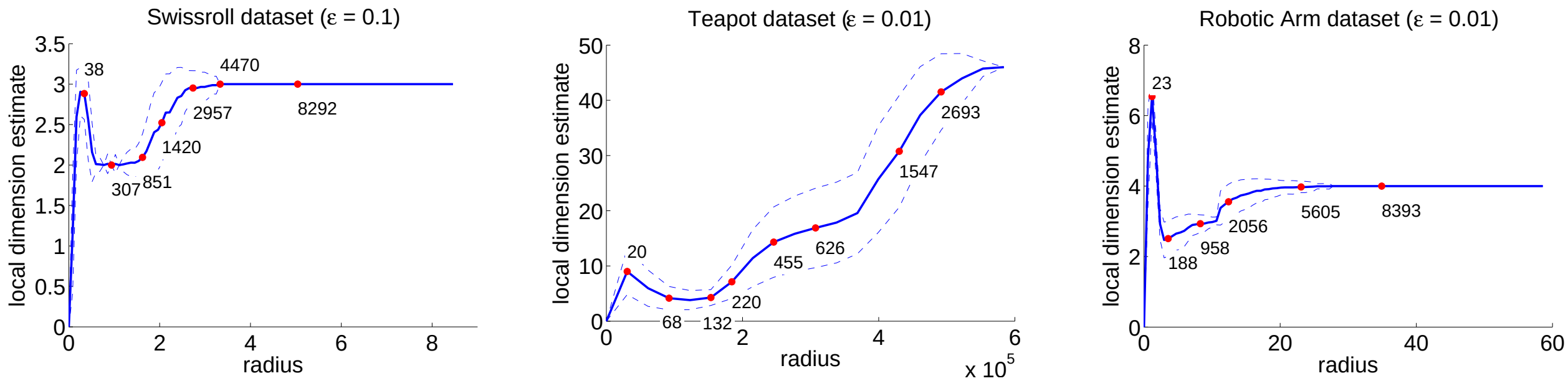

Figure 2: Local Covariance Dimension Estimates for Various Datasets. The bold line shows the dimension estimate $d(r)$, with dashed lines giving standard deviations over the different balls for each radius. The numeric annotations are average numbers of datapoints falling in balls of the specified radius. Left: Noisy swissroll (ambient dimension 3). Middle: Rotating teapot dataset (ambient dimension 1500). Right: Sensors on a robotic arm (ambient dimension 12).

The leftmost figure shows dimensionality estimates for a noisy version of the ever-popular "swiss roll". In small neighborhoods, it is noise that dominates, and thus the data appear full-dimensional. In larger neighborhoods, the two-dimensional structure emerges: notice that the neighborhoods have very large numbers of points, so that we can feel very confident about the estimate of the local covariances. In even larger neighborhoods, we capture a significant chunk of the swiss roll and again revert to three dimensions.

The middle figure is for a data set consisting of images of a rotating teapot, each $30 \times 50$ pixels in size. Thus the ambient dimension is 1500, although the points lie close to a one-dimensional manifold (a circle describing the rotation). There is clear low-dimensional structure at a small scale, although in the figure, these $d(r)$ values seem to be 3 or 4 rather than 1.

The figure on the right is for a data set of noisy measurements from 12 sensors placed on a robotic arm with two joints. Thus the ambient dimension is 12, but there are only two underlying degrees of freedom.

# 3 SPATIAL PARTITION TREES

Spatial partition trees conform to a simple template:

**Procedure** `PartitionTree`($dataset\ A \subset \mathcal{X}$)

**if** $|A| \leq MinSize$ **then**
  **return** leaf
**else**
  $(A_{\text{left}}, A_{\text{right}}) \leftarrow SplitAccordingToSomeRule(A)$
  $LeftTree \leftarrow PartitionTree(A_{\text{left}})$
  $RightTree \leftarrow PartitionTree(A_{\text{right}})$
**return** $(LeftTree, RightTree)$

Different types of trees are distinguished by their splitting criteria. Here are some common varieties:

- **Dyadic tree:** Pick a coordinate direction and splits the data at the midpoint along that direction. One generally cycles through all the coordinates as one moves down the tree.
- **$k$-D tree:** Pick a coordinate direction and splits the data at the median along that direction. One often chooses the coordinate with largest spread.
- **Random Projection (RP) tree:** Split the data at the median along a random direction chosen from the surface of the unit sphere.
- **Principal Direction (PD or PCA) tree:** Split at the median along the principal eigenvector of the covariance matrix.
- **Two Means (2M) tree:** Pick the direction spanned by the centroids of the 2-means solution, and split the data as per the cluster assignment.

## 3.1 NOTIONS OF DIAMETER

The generalization behavior of a spatial partitioning has traditionally been analyzed in terms of the physical diameter of the individual cells (see, for instance, [DGL96, SN06]). But this kind of diameter is hard to analyze for general convex cells. Instead we consider more flexible notions that measure the diameter of *data within the cell*. It has recently been shown that such measures are sufficient for giving generalization bounds (see [Kpo09] for the case of regression).

For any cell $A$, we will use two types of data diameter: the maximum distance between data points in $A$, denoted $\Delta(A)$, and the average interpoint distance among data in $A$, denoted $\Delta_a(A)$ (Figure 3).

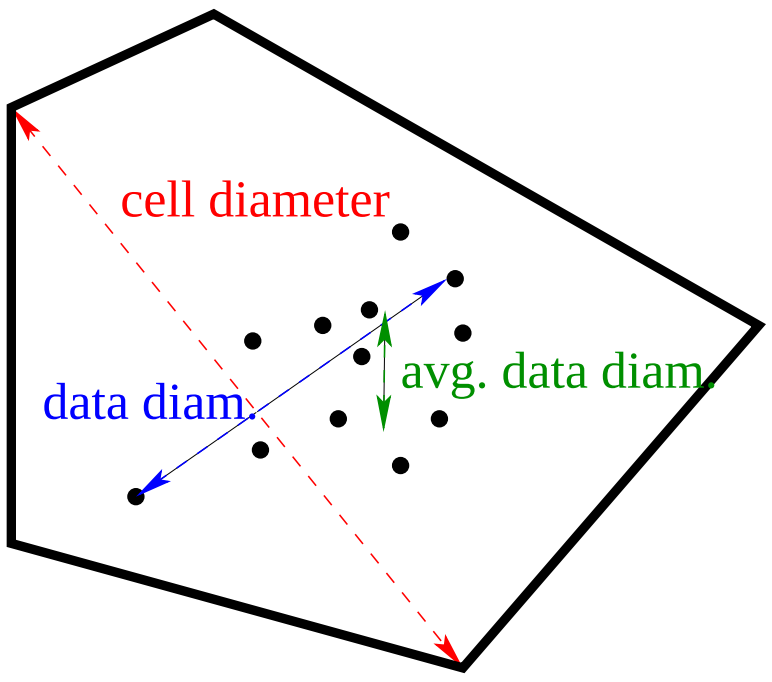

Cell of a Partition Tree

Figure 3: Various Notions of Diameter

## 4 THEORETICAL GUARANTEES

Let $\mathbf{X} = \{X_1, \ldots, X_n\}$ be a data set drawn from $\mathcal{X}$, and let $\mu$ be the empirical distribution that assigns equal weight to each of these points. Consider a partition of $\mathcal{X}$ into a collection of cells $A \in \mathbf{A}$. For each such cell $A$, we can look at its maximum (data) diameter as well as its average (data) diameter; these are, respectively,

$$\Delta(A) \doteq \max_{x,x' \in A \cap \mathbf{X}} \|x - x'\|$$

$$\Delta_a(A) \doteq \frac{1}{(n\mu(A))} \left( \sum_{x,x' \in A \cap \mathbf{X}} \|x - x'\|^2 \right)^{1/2}$$

(for the latter it turns out to be a big convenience to use squared Euclidean distance.) We can also average these quantities all over cells $A \in \mathbf{A}$:

$$\Delta(\mathbf{A}) \doteq \left( \frac{\sum_{A \in \mathbf{A}} \mu(A) \Delta^2(A)}{\sum_{A \in \mathbf{A}} \mu(A)} \right)^{1/2}$$

$$\Delta_a(\mathbf{A}) \doteq \left( \frac{\sum_{A \in \mathbf{A}} \mu(A) \Delta_a^2(A)}{\sum_{A \in \mathbf{A}} \mu(A)} \right)^{1/2}$$

### 4.1 IRREGULAR SPLITTING RULES

This section considers the RPTree, PDtree, and 2Mtree splitting rules. The nonrectangular partitions created by these trees turn out to be adaptive to the local dimension of the data: the decrease in average diameter resulting from a given split depends just on the eigenspectrum of the data in the local neighborhood, irrespective of the ambient dimension.

For the analysis, we consider a slight variant of these trees, in which an alternative type of split is used whenever the data in the cell has outliers (here, points that are much farther away from the mean than the typical distance-from-mean).

**Procedure** split(*region* $A \subset \mathcal{X}$)

**if** $\Delta^2(A) \geq c \cdot \Delta_a^2(A)$ **then**
  //SPLIT BY DISTANCE: remove outliers.
  $A_{\text{left}} \leftarrow \{x \in A, \|x - \text{mean}(A)\| \leq \text{median}\{\|z - \text{mean}(A)\| : z \in \mathbf{X} \cap A\}\}$;
**else**
  //SPLIT BY PROJECTION: no outliers.
  Choose a unit direction $v \in \mathbb{R}^D$ and a threshold $t \in \mathbb{R}$. $A_{\text{left}} \leftarrow \{x \in A, x \cdot v \leq t\}$;
$A_{\text{right}} \leftarrow A \setminus A_{\text{left}}$;

The *distance split* is common to all three rules, and serves to remove outliers. It is guaranteed to reduce maximum data diameter by a constant fraction:

**Proposition 1** (Lemma 12 of [DF08]). *Suppose* $\Delta^2(A) > c \cdot \Delta_a^2(A)$, *so that* $A$ *is split by* distance *under any instantiation of procedure* split. *Let* $\mathbf{A} = \{A_1, A_2\}$ *be the resulting split. We have*

$$\Delta^2(\mathbf{A}) \leq \left( \frac{1}{2} + \frac{2}{c} \right) \Delta^2(A).$$

We consider the three instantiations of procedure split in the following three sections, and we bound the decrease in diameter after a single split in terms of the local spectrum of the data.

#### 4.1.1 RPtree

For RPtree, the direction $v$ is picked randomly, and the threshold $t$ is the median of the projected data.

The diameter decrease after a split depends just on the parameter $d$ of the local covariance dimension, for $\epsilon$ sufficiently small.

**Proposition 2** (Theorem 4 of [DF08]). *There exist constants* $0 < c_1, c_2 < 1$ *with the following property. Suppose* $\Delta^2(A) \leq c \cdot \Delta_a^2(A)$, *so that* $A$ *is split by* projection *into* $\mathbf{A} = \{A_1, A_2\}$ *using the RPtree split. If* $A \cap \mathbf{X}$ *has covariance dimension* $(d, c_1)$, *then*

$$\mathbb{E}\left[\Delta_a^2(\mathbf{A})\right] < (1 - c_2/d)\Delta_a^2(A),$$

*where the expectation is over the choice of direction.*

#### 4.1.2 PDtree

For PDtree, the direction $v$ is chosen to be the principal eigenvector of the covariance matrix of the data, and the threshold $t$ is the median of the projected data.

The diameter decrease after a split depends on the local spectrum of the data. Let $A$ be the current cell being split, and suppose the covariance matrix of the data in $A$ has eigenvalues $\lambda_1 \geq \cdots \geq \lambda_D$. If the co-variance dimension of $A$ is $(d, \epsilon)$, define

$$k \doteq \frac{1}{\lambda_1}\sum_{i=1}^{d}\lambda_i, \quad (1)$$

By definition, $k \leq d$.

The diameter decrease after the split depends on $k^2$, the worst case being when the data distribution in the cell has heavy tails (example omitted for want of space). In the absence of heavy tails (condition (2)), we obtain a faster diameter decrease rate that depends just on $k$. This condition holds for any logconcave distribution (such as a Gaussian or uniform distribution), for instance. The decrease rate of $k$ could be much better than $d$ in situations where the first eigenvalue is dominant; and thus in such situations PD trees could do a lot better than RP trees.

**Proposition 3.** *There exist constants $0 < c_1, c_2 < 1$ with the following property. Suppose $\Delta^2(A) \leq c \cdot \Delta_a^2(A)$, so that $A$ is split by* projection *into $\mathbf{A} = \{A_1, A_2\}$ using the PDtree split. If $A \cap \mathbf{X}$ has covariance dimension $(d, c_1)$, then*

$$\Delta_a^2(\mathbf{A}) < (1 - c_2/k^2)\Delta_a^2(A),$$

*where $k$ is as defined in (1).*

*If in addition the empirical distribution on $A \cap \mathbf{X}$ satisfies (for any $s \in \mathbb{R}$ and some $c_0 \geq 1$)*

$$\mathbb{E}_A[(X \cdot v - s)^2] \leq c_0 (\mathbb{E}_A[X \cdot v - s])^2 \quad (2)$$

*we obtain a faster decrease where*

$$\Delta_a^2(\mathbf{A}) < (1 - c_2/k)\Delta_a^2(A).$$

*Proof.* The argument is based on the following fact which holds for any bi-partiton $\mathbf{A} = \{A_1, A_2\}$ of $A$ (see lemma 15 of [DF08]):

$$\begin{aligned}&\Delta_a^2(A) - \Delta_a^2(\mathbf{A})\\ &= 2\mu(A_1) \cdot \mu(A_2) \left\|\text{mean}(A_1) - \text{mean}(A_2)\right\|^2 \quad (3)\end{aligned}$$

We start with the first part of the statement with no assumption on the data distribution. Let $\widetilde{x} \in \mathbb{R}$ be the projection of $x \in A \cap \mathbf{X}$ to the principal direction. WLOG assume that the median on the principal direction is 0. Notice that

$$\begin{aligned}\|\text{mean}(A_1) - \text{mean}(A_2)\| &\geq \mathbb{E}[\widetilde{x}|\widetilde{x} > 0] - \mathbb{E}[\widetilde{x}|\widetilde{x} \leq 0]\\ &\geq \max\{\mathbb{E}[\widetilde{x}|\widetilde{x} > 0], -\mathbb{E}[\widetilde{x}|\widetilde{x} \leq 0]\}\end{aligned}$$

where the expectation is over $x$ chosen uniformly at random from $A \cap \mathbf{X}$. The claim is therefore shown by bounding the r.h.s below by $O(\Delta_a(A)/k$ and applying equation (3).

We have $\mathbb{E}\left[\widetilde{x}^2\right] \geq \lambda_1$, so either $\mathbb{E}\left[\widetilde{x}^2|\widetilde{x} > 0\right]$ or $\mathbb{E}\left[\widetilde{x}^2|\widetilde{x} \leq 0\right]$ is greater than $\lambda_1$. Assume WLOG that it is the former. Let $\widetilde{m} = \max\{\widetilde{x} > 0\}$. We have that

$$\lambda_1 \leq \mathbb{E}\left[\widetilde{x}^2|\widetilde{x} > 0\right] \leq \mathbb{E}\left[\widetilde{x}|\widetilde{x} > 0\right]\widetilde{m},$$

and since $\widetilde{m}^2 \leq c\Delta_a^2(A)$, we get

$$\mathbb{E}\left[\widetilde{x}|\widetilde{x} > 0\right] \geq \frac{\lambda_1}{\Delta_a(A)\sqrt{c}}.$$

Now, by the assumption on covariance dimension,

$$\lambda_1 = \frac{\sum_{i=1}^{d}\lambda_i}{k} \geq (1 - c_1)\frac{\sum_{i=1}^{D}\lambda_i}{k} = (1 - c_1)\frac{\Delta_a^2(A)}{2k}.$$

We therefore have (for appropriate choice of $c_1$) that $\mathbb{E}[\widetilde{x}|\widetilde{x} > 0] \geq \Delta_a(A)/4k\sqrt{c}$, which concludes the argument for the first part.

For the second part, assumption (2) yields

$$\begin{aligned}\mathbb{E}[\widetilde{x}|\widetilde{x} > 0] - \mathbb{E}[\widetilde{x}|\widetilde{x} \leq 0] &= 2\mathbb{E}|\widetilde{x}| \geq 2\sqrt{\frac{\mathbb{E}|\widetilde{x}|^2}{c_0}}\\ &\geq 2\sqrt{\frac{\lambda_1}{c_0}} = 2\sqrt{\frac{\Delta_a^2(A)}{4c_0 k}}.\end{aligned}$$

We finish up by appealing to equation (3). □

#### 4.1.3 2Mtree

For 2Mtree, the direction $v = \text{mean}(A_1) - \text{mean}(A_2)$ where $A = \{A_1, A_2\}$ is the bisection of $A$ that minimizes the 2-means cost. The threshold $t$ is the half point between the two means.

The 2-means cost can be written as

$$\sum_{i \in [2]}\sum_{x \in A_i \cap \mathbf{X}} \|x - \text{mean}(A_i)\|^2 = \frac{n}{2}\Delta_a^2(\mathbf{A}).$$

Thus, the 2Mtree (assuming an exact solver) minimizes $\Delta_a^2(\mathbf{A})$. In other words, it decreases diameter at least as fast as RPtree and PDtree. Note however that, since these are greedy procedures, the decrease in diameter over multiple levels may not be superior to the decrease attained with the other procedures.

**Proposition 4.** *Suppose $\Delta^2(A) \leq c \cdot \Delta_a^2(A)$, so that $A$ is split by* projection *into $\mathbf{A} = \{A_1, A_2\}$ using the 2Mtree split. There exists constants $0 < c_1, c_2 < 1$ with the following property. Assume $A \cap \mathbf{X}$ has covariance dimension $(d, c_1)$. We then have*

$$\Delta_a^2(\mathbf{A}) < (1 - c_2/d')\Delta_a^2(A),$$

*where $d' \leq \min\{d, k^2\}$ for general distributions, and $d'$ is at most $k$ for distributions satisfying (2).*

#### 4.1.4 Diameter Decrease Over Multiple Levels

The diameter decrease parameters $d, k^2, k, d'$ in propositions 2, 3, 4 above are a function of the covariance dimension of the data in the cell $A$ being split. The covariance dimensions of the cells may vary over the course of the splits implying that the decrease rates may vary. However, we can bound the overall diameter decrease rate over multiple levels of the tree in terms of the worst case rate attained over levels.

**Proposition 5** (Diameter decrease over multiple levels)**.** *Suppose a partition tree is built by calling* `split` *recursively (under any instantiation). Assume furthermore that every node $A \subset \mathcal{X}$ of the tree satisfies the following: let $\mathbf{A} = \{A_1, A_2\}$ represent the child nodes of $A$, we have for some constants $0 < c_1, c_2 < 1$ and $\kappa \leq D$ that*

*(i) If $A$ is split by distance, $\Delta^2(\mathbf{A}) < c_1 \Delta^2(A)$.*

*(ii) If $A$ is split by projection, $\mathbb{E}\left[\Delta_a^2(\mathbf{A})\right] < (1 - c_2/\kappa)\Delta_a^2(A)$.*

*Then, there exists a constant $C$ such that the following holds: let $\mathbf{A}_l$ be the partition of $\mathcal{X}$ defined by the nodes at level $l$, we have*

$$\mathbb{E}\left[\Delta_a^2(\mathbf{A}_l)\right] \leq \mathbb{E}\left[\Delta^2(\mathbf{A}_l)\right] \leq \frac{1}{2^{\lfloor l/C\kappa \rfloor}} \Delta^2(\mathcal{X})$$

.

*In all the above, the expectation is over the randomness in the algorithm for $\mathbf{X}$ fixed.*

*Proof.* Fix $\mathbf{X}$. Consider the r.v. $X$ drawn uniformly from $\mathbf{X}$. Let the r.v.s $A_i = A_i(X), i = 0 \cdots l$ denote the cell to which $X$ belongs at level $i$ in the tree. Define $I(A_i) \doteq \mathbb{1}\left\{\Delta^2(A_i) \leq c\Delta_a^2(A_i)\right\}$.

Let $\mathbf{A}_l$ be the partition of $\mathcal{X}$ defined by the nodes at level $l$, we'll first show that $\mathbb{E}\left[\Delta^2(\mathbf{A}_l)\right] \leq \frac{1}{2}\Delta^2(\mathcal{X})$ for $l = C\kappa$ for some constant $C$. We point out that $\mathbb{E}\left[\Delta^2(\mathbf{A}_l)\right] = \mathbb{E}\left[\Delta^2(A_l)\right]$ where the last expectation is over the randomness in the algorithm and the choice of $X$.

To bound $\mathbb{E}\left[\Delta^2(A_l)\right]$, note that one of the following events must hold:

(a) $\exists\ 0 \leq i_1 < \cdots < i_m < l,\ m \geq \frac{l}{2}, I\left(A_{i_j}\right) = 0$

(b) $\exists\ 0 \leq i_1 < \cdots < i_m < l,\ m \geq \frac{l}{2}, I\left(A_{i_j}\right) = 1$

Let's first condition on event (a). We have

$$\begin{aligned}\mathbb{E}\left[\Delta^2(A_l)\right] &\leq \mathbb{E}\left[\Delta^2(A_{i_m+1})\right] \\ &= \mathbb{E}\left[\mathbb{E}\left[\Delta^2(A_{i_m+1})\,|A_{i_m}\right]\right],\end{aligned}$$

and since by the assumption, $\mathbb{E}\left[\Delta^2(A_{i_m+1})\,|A_{i_m}\right] \leq c_1\Delta^2(A_{i_m})$ we get that $\mathbb{E}\left[\Delta^2(A_l)\right] \leq c_1\mathbb{E}\left[\Delta^2(A_{i_m})\right]$. Applying the same argument recursively on $i_j$, $j = m, (m-1), \ldots, 1$, we obtain

$$\mathbb{E}\left[\Delta^2(A_l)\right] \leq c_1^m \cdot \mathbb{E}\left[\Delta^2(A_{i_1})\right] \leq c_1^{l/2}\Delta^2(\mathcal{X}).$$

Now condition on event (b). Using the fact that $\mathbb{E}\left[\Delta_a^2(A_i)\right]$ is non-increasing in $i$ (see [DF08]), we can apply a similar recursive argument as above to obtain that $\mathbb{E}\left[\Delta_a^2(A_{i_m})\right] \leq (1 - c_2/\kappa)^{m-1}\mathbb{E}\left[\Delta_a^2(A_{i_1})\right]$. It follows that

$$\begin{aligned}\mathbb{E}\left[\Delta^2(A_l)\right] &\leq \mathbb{E}\left[\Delta^2(A_m)\right] \leq c\mathbb{E}\left[\Delta_a^2(A_m)\right] \\ &\leq c\left(1 - \frac{c_2}{\kappa}\right)^{l/2-1}\Delta^2(\mathcal{X}).\end{aligned}$$

Thus, in either case we have

$$\mathbb{E}\left[\Delta^2(A_l)\right] \leq \max\left\{c_1^{l/2},\ c\left(1 - c_2/\kappa\right)^{l/2-1}\right\} \cdot \Delta^2(\mathcal{X})$$

and we can verify that there exists $C$ such that the r.h.s above is at most $\frac{1}{2}\Delta^2(\mathcal{X})$ for $l \leq C\kappa$. Thus, we can repeat the argument over every $C\kappa$ levels to obtain the statement of the proposition. □

So if every split decreases average diameter at a rate controlled by $\kappa$ as defined above, then it takes at most $O(\kappa \log(1/\varepsilon))$ levels to decrease average diameter down to an $\varepsilon$ fraction of the original diameter of the data. Combined with propositions 2, 3, 4, we see that the three rules considered will decrease diameter at a fast rate whenever the covariance dimensions in local regions are small.

### 4.2 AXIS PARALLEL SPLITTING RULES

It was shown in [DF08] that axis-parallel splitting rules do not always adapt to data that is intrinsically low-dimensional. They exhibit a data set in $\mathbb{R}^D$ that has low Assouad dimension $O(\log D)$, and where $k$-d trees (and also, it can be shown, dyadic trees) require $D$ levels to halve the data diameter.

The adaptivity of axis-parallel rules to covariance dimension is unclear. But they *are* guaranteed to decrease diameter at a rate depending on $D$. The following result states that it takes at most $O(D(\log D)\log(1/\varepsilon))$ levels to decrease average diameter to an $\varepsilon$ fraction of the original data diameter.

**Proposition 6.** *Suppose a partition tree is built using either $k$-d tree or dyadic tree by cycling through the coordinates. Let $\mathbf{A}_l$ be the partition of $\mathcal{X}$ defined by the nodes at level $l$. Then we have*

$$\Delta_a^2(\mathbf{A}_l) \leq \Delta^2(\mathbf{A}_l) \leq \frac{D}{2^{\lfloor l/D \rfloor}}\Delta^2(\mathcal{X}).$$

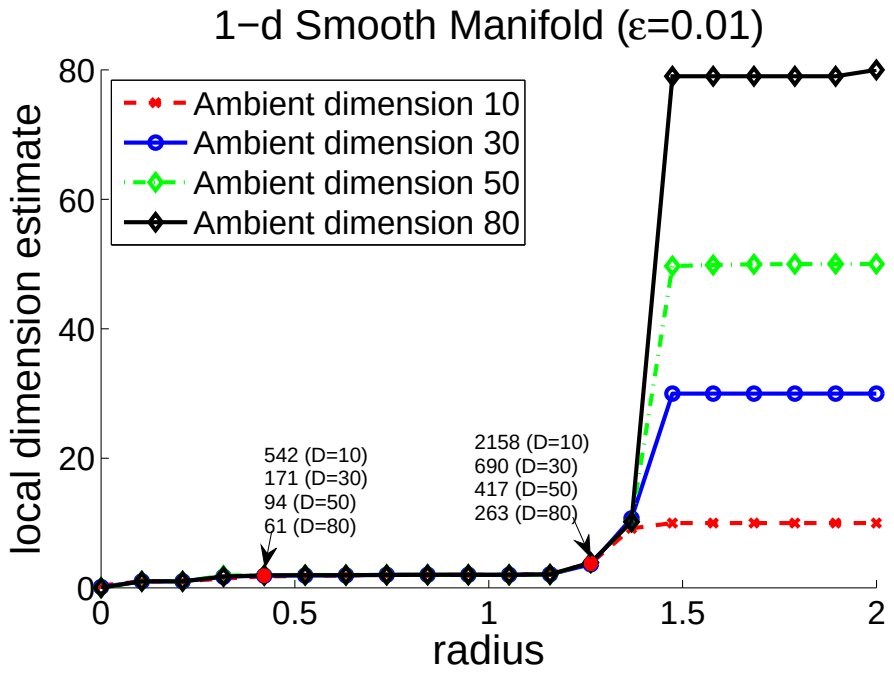

Figure 4: Local Covariance Dimension Estimate of the 1-d Manifold

# 5 EXPERIMENTS

To highlight the adaptivity of the trees considered, we first run experiments on a synthetic dataset where we can control the intrinsic and the ambient dimensions. We vary the ambient dimension while keeping the intrinsic dimension small, and monitor the rate at which the diameter decreases. Next, we compare the performance of these trees on common learning tasks using real-world datasets.

We implement the trees as follows: dyadic trees – fix a permutation and cycle through the coordinates, $k$-D trees – determine the spread over each coordinate by computing the coordinate vise diameter and picking the coordinate with maximum diameter, RP trees – pick the direction that results in the largest diameter decrease from a bag of 20 random directions, PD trees – pick the principal direction in accordance to the data falling in each node of the tree, 2M trees – solve 2-means via the Lloyd's method and pick the direction spanned by the centroids of the 2-means solution.

## 5.1 SYNTHETIC DATASET: SMOOTH 1-D MANIFOLD

The synthetic dataset used is a continuous, 1 dimensional manifold obtained via the sinusoidal basis as follows. We sample $20,000$ points uniformly at random from the interval $[0, 2\pi]$, and for each point $t$, we apply the smooth map $M : t \mapsto \sqrt{\frac{2}{D}}\left(\sin(t), \cos(t), \sin(2t), \cos(2t)\ldots, \sin(\frac{Dt}{2}), \cos(\frac{Dt}{2})\right)$.

Figure 4 shows the local covariance dimension estimate for this 1-d manifold (embedded in ambient space of dimension $D = 10$, 30, 50 and 80).

What behavior should we expect from adaptive trees? Taking a cue from proposition 5, we expect the diameter to decrease initially at a rate controlled by the ambient space, since the covariance dimension is high in large regions (cf. figure 4). When we get to smaller regions, the diameter should decrease down the tree by a factor which is controlled solely by the local covariance dimension.

We therefore plot the log of the diameter as a function of the tree depth, for each value of $D$, and monitor the slope of this function (see figure 5). The slope, which corresponds to the diameter decrease rate, should eventually coincide in small regions for all values of $D$, provided the partition tree is adaptive to the intrinsic dimensionality of the data.

In figure 5, notice that slopes for $k$-D, RP, PD and 2M trees start to converge after about depth 7. This indicates that the ambient dimension has negligible effect on the behavior of these trees. Dyadic trees, however, don't perform as well.

## 5.2 REALWORLD DATASETS

We now compare the performance of different spatial trees for typical learning tasks on some real-world datasets. To exhibit a wide range of applicability, we choose the 'digit 1' cluster from the MNIST OCR dataset of handwritten digits, 'love' cluster from Australian Sign Language time-series dataset from UCI Machine Learning Repository [Kad02], and 'aw' phoneme from MFCC TIMIT dataset. We also use the rotating teapot and the robotic arm datasets to evaluate regression performance (cf. section 2.2).

Experiments are set as follows. For each dataset, we first estimate local covariance dimensions (as discussed in section 2.2). We then perform three types of experiments: vector quantization, nearest neighbor search and regression. For each experiment, we do a 10-fold cross validation. For each fold, we use the training data to build the partition tree, and compute the quantization error, the closest neighbor and the regression error over the induced partition at each level.

Overall, we see that 2M trees and PD trees perform consistently better in the three types of experiments.

**Vector Quantization:** Figure 6 middle row shows the relative quantization errors for the trees. The PD and 2M trees produce better quantization results than other trees, followed by RP trees.

**Near Neighbor Search:** Figure 6 bottom row shows the result of a near neighbor search. The plot shows the order of nearest neighbor found normalized by the dataset size (percentile order). The annotated numbers show the ratio of the distance between the query point and the discovered neighbor to the distance between the query point and its true nearest neighbor. This helps in gauging the quality of the found neighbor in terms of distances.

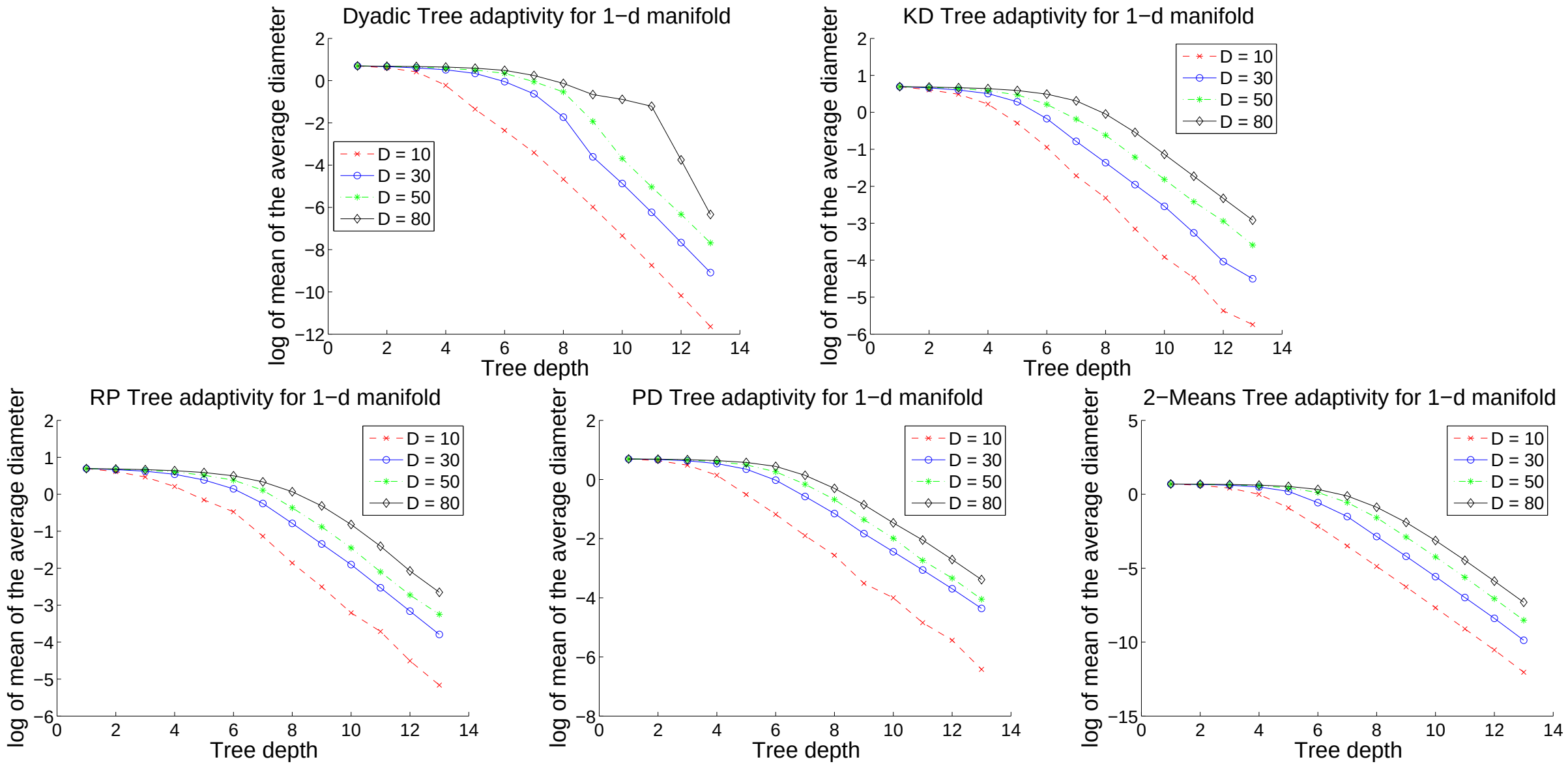

Figure 5: Adaptivity Plots for Various Spatial Trees on the Synthetic Dataset. Note that the *slope* of the plotted curve shows the decrease rate. Parallel lines highlight that the diameter decrease rates eventually become *independent* of the ambient dimension adapting to the low dimensional intrinsic structure of the manifold.

As before, 2M and PD trees consistently yield better near neighbors to the query point. We should remark that the apparent good results of dyadic trees on the ASL dataset (middle row, middle column) should be taken in context with the number of datapoints falling in a particular node. For dyadic, trees it is common to have unbalanced splits resulting in high number of datapoints falling in an individual cell. This significantly increases the chance of finding a close neighbor but also increases its computational cost.

**Regression:** For regression on the teapot dataset, we predict the value of the rotation angle which ranges between 0 and 180 degrees. Recall that this dataset lies on a 1 dimensional manifold in $\mathbb{R}^{1500}$. From known regression results, we expect the most adaptive trees to yield lower regression errors. Here, these are the 2M tree and PD tree, followed by RP tree.

The robotic arm dataset is a noisy two dimensional manifold in $\mathbb{R}^{12}$. From our covariance dimension estimation (see figure 2 right), the data resides close to a 4 dimensional linear space. Not surprisingly all trees perform comparably well, except on arm 2, where dyadic tree does significantly worse.

## Acknowledgements

We are grateful for financial support from the National Science Foundation (under grants IIS-0347646, IIS-0713540, and IIS-0812598) and from the Engineering Institute at the Los Alamos National Laboratory.

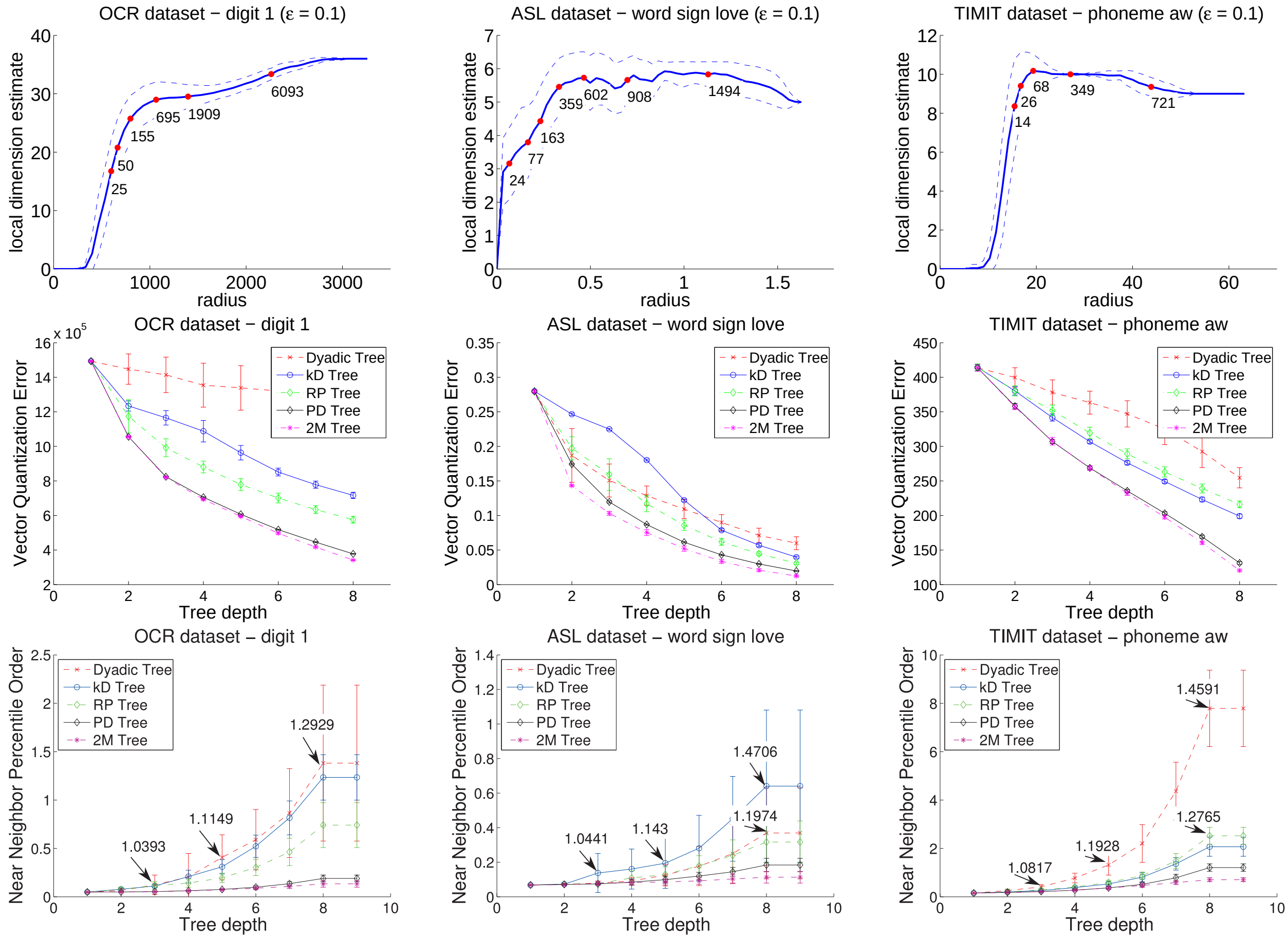

Figure 6: Top: Local covariance dimension estimates for real-world datasets, with confidence bands (dashed lines) and average number of points falling in balls of specified radius (numerical annotations). Middle: Average vector quantization error induced by different spatial trees on the datasets. Bottom: Results for near neighbor query. Annotated numbers show the average ratio of the distance between the query point and the found neighbor to the distance between the query point and the true nearest neighbor.

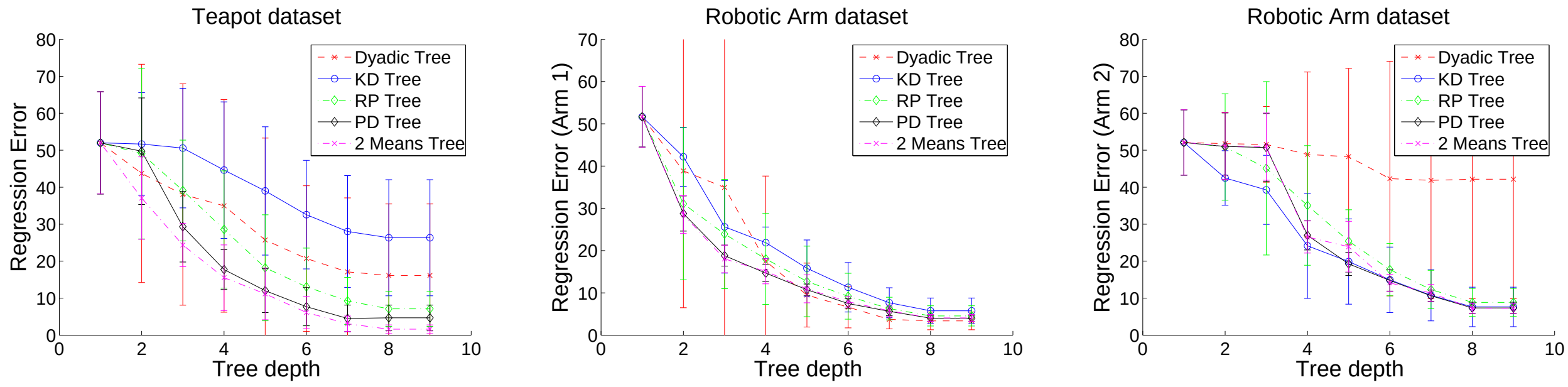

Figure 7: Relative Performance of Spatial Trees on a Regression Task. $\ell_2$ error is being computed. Left: Teapot dataset, predicting the rotation angle. Middle, Right: Robotic arm dataset, predicting the angular positions of arm one and arm two.